%% file: 0_Main.tex
\documentclass[10pt,twocolumn]{article}

\usepackage{times}
\usepackage{graphicx}
\usepackage{amsmath}
\usepackage{amssymb}
\usepackage[export]{adjustbox}
\usepackage{makecell}
\usepackage{authblk}
\usepackage{placeins}
\begin{document}

\title{IJCB 2022 Mobile Behavioral Biometrics Competition (MobileB2C)}
\author{\small Giuseppe Stragapede$^{1, }$*, Ruben Vera-Rodriguez$^1$, Ruben Tolosana$^1$, Aythami Morales$^1$, Julian Fierrez$^1$, Javier Ortega-Garcia$^1$, \\ Sanka Rasnayaka$^2$, Sachith Seneviratne$^3$, Vipula Dissanayake$^4$, Jonathan Liebers$^5$, Ashhadul Islam$^6$, Samir Brahim Belhaouari$^6$,\\ Sumaiya Ahmad$^7$, Suraiya Jabin$^7$}

\date{}

\affil{\small $^1$Biometrics and Data Pattern Analytics Lab, UAM, Madrid, Spain, $^2$National University of Singapore, Singapore,\\ $^3$University of Melbourne, Melbourne, Australia, $^4$University of Auckland, Auckland, New Zealand,\\ $^5$University of Duisburg-Essen, Duisburg, Germany, $^6$Hamad Bin Khalifa University, ar-Rayyan, Qatar,\\ $^7$Jamia Millia Islamia, New Delhi, India} 

\affil{\small{giuseppe.stragapede@uam.es (corresponding author)}} 


\maketitle

\begin{abstract}
This paper describes the experimental framework and results of the IJCB 2022 Mobile Behavioral Biometrics Competition (MobileB2C). The aim of MobileB2C is benchmarking mobile user authentication systems based on behavioral biometric traits transparently acquired by mobile devices during ordinary Human-Computer Interaction (HCI), using a novel public database, BehavePassDB\footnote{https://github.com/BiDAlab/MobileB2C\_BehavePassDB}, and a standard experimental protocol. The competition is divided into four tasks corresponding to typical user activities: keystroke, text reading, gallery swiping, and tapping. The data are composed of touchscreen data and several background sensor data simultaneously acquired. “Random” (different users with different devices) and “skilled” (different user on the same device attempting to imitate the legitimate one) impostor scenarios are considered. The results achieved by the participants show the feasibility of user authentication through behavioral biometrics, although this proves to be a non-trivial challenge. MobileB2C will be established as an on-going competition \footnote{https://sites.google.com/view/mobileb2c/}.
\end{abstract}

\section{Introduction}
\label{sec:Introduction}
\input{1_Introduction}

\section{The Data: BehavePassDB}
\label{sec:The_Data_BehavePassDB}
\input{2_BehavePassDB}

\section{Competition Set Up}
\label{sec:Competition_Set_Up}
\input{3_Competition_Set_Up}

\section{Description of the Evaluated Systems}
\label{sec:Description_of_the_Evaluated_Systems}
\input{4_Description_of_the_Evaluated_Systems}

\section{Experimental Results}
\label{sec:Experimental_Results}
\input{5_Experimental_Results}

\section{Conclusions}
\label{sec:Conclusions}
\input{6_Conclusions}

\section*{Acknowledgments}

This project has received funding from the European Union’s Horizon 2020 research and innovation programme under the Marie Skłodowska-Curie grant agreement No 860315, and from Orange Labs. R. Vera-Rodriguez and R. Tolosana are also supported by INTER-ACTION (PID2021-126521OB-I00 MICINN/FEDER). 

{\small
\bibliographystyle{ieee}
\bibliography{0_Main}
}

\end{document}

%% file: 1_Introduction.tex

All biometric data that enable or contribute to differentiating between individuals throughout the \textit{way} they perform activities are labelled as \textit{behavioral}. Behavioral biometrics are a broad category including voice \cite{voice}, gait \cite{delgadosantos2022exploring}, keystroke \cite{9539873}, or handwritten signature \cite{tolosana2022svc}, among others. They are recently gaining more attention from academia and industry as a new solution for the problem of mobile user authentication \cite{stragapede2022prl}. In fact, mobile devices are provided with numerous sensors, i.e., touchscreen, motion sensors, etc., performing continuous measurements in the form of low-dimensional temporal signals, which contain a large amount of user information \cite{delgadosantos2021survey}. Nonetheless, as of today, \textit{physiological}\footnote{All biological characteristics that allow to identify an individual are defined as \textit{physiological} (face, fingerprint, iris, etc.).} biometrics are the most popular traits employed for biometric mobile user authentication \cite{WANG2020107118}. However, just like security systems based on passwords, they are not designed to provide extended protection over the entire device usage session. Once an adversary has overcome the entry-point security mechanism such as fingerprint verification, they will not be locked out as long as the device is active. Thus, private information can be exposed for a prolonged period of time \cite{Perera2017}. A continuous face verification process also appears impractical in the light of the limitations of mobile hardware. On the contrary, behavioral biometrics are suitable for Continuous Authentication (CA), a security paradigm based on the continued acquisition and verification of user biometric traits \textit{passively}, i.e., without a specific authentication activity required to the user \cite{7503170, stragapede2022prl}. 
\par To contribute to the development of continuous authentication based on mobile behavioral biometrics, we propose the Mobile Behavioral Biometrics Competition (MobileB2C), described in the remainder of the paper.

\subsection{Scope of the Competition}
\label{subsec:Scope_of_the_Competition}
In the field of mobile behavioral biometrics, a challenge for the research community is given by the scarcity of public databases, and by the fact that the recent, most promising studies in the field are often very heterogeneous \cite{Abuhamad, Acien2020b, 9539873, 2020_CDS_HCIsmart_Acien, Deb2019, melzi2022ecg, stragapede2022prl}. Consequently, it would be difficult to reach a global and significant conclusion from the comparison of such systems, given the different approaches, scopes and the usage of self-collected non-public databases.
\par In this scenario, the aim of the Mobile Behavioral Biometrics Competition (MobileB2C) is to benchmark mobile behavioral biometric authentication systems that are transparently acquired by mobile devices during ordinary Human-Computer Interaction (HCI), using a recently proposed database, BehavePassDB (Sec. \ref{sec:The_Data_BehavePassDB}), and a standard experimental protocol (Sec. \ref{subsec:Experimental_Protocol}). 
\par MobileB2C is based on four different tasks, corresponding to common mobile device use cases: \textit{(i)} texting, \textit{(ii)} text reading, \textit{(iii)} gallery swiping, \textit{(iv)} tapping. In each of the tasks, the touchscreen information is recorded along with the data of five background sensors: \textit{(a)} accelerometer, \textit{(b)} gyroscope, \textit{(c)} magnetometer, \textit{(d)} linear accelerometer, \textit{(e)} gravity sensor.

\subsection{Competition Novel Aspects}
\label{subsec:Competition_Novel_Aspects}
Currently, behavioral biometrics cannot guarantee the same authentication performance level as \textit{physiological} biometrics yet. Nevertheless, the attention of the biometrics community for Behavioral Biometrics for Continuous Authentication (BBCA) is rising \cite{7503170}. We outline two potential application scenarios:
\begin{itemize}
\itemsep-.5em 
    \item Second factor channel in a two-factor authentication (2FA) \cite{8998358} in security-wise critical contexts such as during the usage of a bank or healthcare application; users can be redirected to an \textit{entry-point} authentication mechanism in case that the matching of the continuously acquired biometric samples with pre-acquired enrolment ones returns a negative response. \textbf{In this case, every user would be using their own mobile device.} 
    \item The main security mechanisms in the circumstance of device theft, in which the \textbf{impostor and the genuine user data originate from the same device} \cite{rasmussen15NDSS}. 
\end{itemize}	
In the first case, the features recognized by the authentication system could be – to a certain extent - related to the device rather than to the user biometric traits. In fact, a device bias could be present due to sensor differences and calibration imperfections across devices \cite{das2015exploring, Neverova2016}. 
\par In the second case, the authentication system must be free of any device bias and able to extract and distinguish among purely biometric information only. This condition leads to additional challenges during the system development stage, if only one device is considered, as the system performance could be affected when deployed on different devices.
\par In this context, the goal of MobileB2C is carrying out a benchmark evaluation of the latest state-of-the-art mobile behavioral biometrics authentication technology using a large-scale public database, BehavePassDB, which encompasses both traditional “random impostor” scenario (different users with different devices), but also the challenging “skilled impostor” scenario (different users on the same device attempting to imitate the device owner).

%% file: 2_BehavePassDB.tex
MobileB2C is based on the recently proposed BehavePassDB \cite{BehavePassDB}, containing user data designed to represent typical circumstances of mobile HCI. The data collection process involved four sessions, acquired 24 hours apart to take into consideration intra-subject biometric variability. The subjects installed an Android mobile app on their personal smartphone and completed eight tasks without any supervision.
The tasks are free-text keystroke, reading a text, swiping gallery images, tapping on the screen, handwritten signature, and picking up the smartphone, and a reproduction of a security-wise critical app (bank mobile app), that includes a combination of the other tasks. In each of the tasks, up to 15 sensors carry out continuous measurements. They are: accelerometer, linear accelerometer, gyroscope, magnetometer, ambient temperature, proximity, gravity, light, humidity, pressure, GPS, Wi-Fi, Bluetooth, and battery.
\par The BehavePassDB is divided into two subsets: \textit{(i)} a development set, containing 61 users, divided into a training set (51 users), and a validation set (10 users), and \textit{(ii)} an evaluation set, containing different 20 users. The training set includes 4 sessions per user, acquired while each of the users was using their own device (random impostor scenario). In contrast, the validation and evaluation sets also include two additional sessions (leading to 6 sessions in total per acquisition device) in which a different user (skilled impostor) is using the same device as the owner in the attempt to imitate them, allowing both random and skilled impostor scenarios.  
Out of all tasks and sensors available in BehavePassDB, the ones considered in the competition are indicated in Table \ref{tab:table_tasks}.

\begin{table}[ht!]
\caption{The tasks included in the competition with the corresponding available data are displayed below.}
\centering
\label{tab:table_tasks}
\begin{tabular}{c c}
\hline
\textbf{Task} & \textbf{Data} \\
\Xhline{2\arrayrulewidth}
\makecell[l]{1. Texting} & \makecell[l]{Keystroke dynamics,\\ background sensors$^1$} \\
\hline
\makecell[l]{2. Text Reading} & \makecell[l]{Vertical swipe gestures,\\ background sensors} \\
\hline
\makecell[l]{3. Gallery Swiping} & \makecell[l]{Horizontal swipe gestures,\\ background sensors} \\
\hline
\makecell[l]{4. Tapping} & \makecell[l]{Tap gestures,\\ background sensors} \\
\hline
\multicolumn{2}{l}{\makecell[l]{$^1$The background sensors available for each of the tasks\\ are: accelerometer, gravity sensor, gyroscope, linear\\accelerometer, and magnetometer.}}\\
\end{tabular}
\end{table}

\subsection{Data Format}
\label{subsec:Data_Format}
Starting from the acquired raw data, to simplify the data preprocessing and to provide a common starting ground, the time-series data are provided into a different form depending on the modality. Specifically, each acquired sample is structured as follows:
\begin{itemize}
\itemsep-.5em 
    \item[\textit{i)}] Keystroke data:
    \begin{center}
\textit{[timestamp, ascii\_code]
}\end{center}
\item[\textit{ii)}] All other touch data:
    \begin{center}
    \textit{[timestamp, x\_coordinate / screen\_width,\\ y\_coordinate / screen\_height, action\_type]}\\
    \end{center}
    \textit{action\_type} refers to a specific touch event. The value ``0" corresponds to laying the finger, ``1" to lifting the finger, ``2" to moving the finger on the screen.
\item[\textit{iii)}] Background sensor data:
\begin{center}
\textit{[timestamp, x\_coordinate, y\_coordinate, z\_coordinate]
}\end{center}
\end{itemize}
Each of the sets is released in the form of a JSON file.

%% file: 3_Competition_Set_Up.tex
The competition is hosted on CodaLab\footnote{https://codalab.lisn.upsaclay.fr/competitions/3564}, a powerful open-source framework for running scientific competitions and benchmarks \cite{tolosana2022svc}, frequently used in research and education.

\subsection{Experimental Protocol}
\label{subsec:Experimental_Protocol}
\par Each team could choose freely the tasks to focus on based on their expertise in the corresponding modalities (or combination of modalities). In other words, they could make use of the information provided as they like. For example, if they developed a system based on background sensors, they could participate in all tasks, or if the system developed was based on keystroke dynamics, they could participate only in Task 1. The participants were encouraged to exploit the multimodal nature of the available data by considering the fusion of the different biometric modalities available within each task, although this was not mandatory.
\par The data have been released to the participants in two batches at two different times corresponding to the beginning of the two phases of the competition: 
\begin{itemize}
    \item[\textit{(i)}] The initial Validation Phase, for which the development set was provided to the participants. The development set is split in two sets: the training set, indicating in plain user identity codes, and the validation set, including data pseudonymized at session level.
    The validation data were released together with a list of pseudonymized session comparisons for the participants to carry out. 
    \item[\textit{(ii)}] The final Evaluation Phase, whose structure and modality of the release are analogous to the ones of the validation set.
\end{itemize}
A grading program was implemented in Codalab, in order for the participant to submit their session comparison scores  (Sec. \ref{subsec:Evaluation_Criteria}), and validate their system without sharing any aspect of their code. During the initial Validation Phase, the participants could see their scores displayed on a leader board, while in the final Evaluation Phase, the scores were made public after the end of the competition.

\subsection{Evaluation Criteria}
\label{subsec:Evaluation_Criteria}
\par In both phases of the competition, 2 out of the 4 genuine acquisition sessions were used for user enrollment, while the remaining 2 for user verification, as well as the remaining 2 impostor sessions. Starting from the list of scores submitted, the grading program reconstructs genuine, random impostor, and skilled impostor distributions. A popular metric in the field such as the Area Under the Curve (AUC) was used to evaluate the performance of different systems in each of the impostor scenarios (“AUC Random Case”, “AUC Skilled Case”, “AUC Mixed Case”), and the results were displayed in a dedicated leaderboard on Codalab. The final ranking is based on the “AUC Mixed Case” only, which includes both types of impostor distributions. Finally, at the end of the competition, there is one winner per Task, based on the Evaluation Phase results.

%% file: 4_Description_of_the_Evaluated_Systems.tex
A total of 24 participants initially registered in MobileB2C. However, only 4 of them eventually submitted their scores with a total of 9 different behavioral biometrics authentication systems. Next, we describe briefly the systems provided by each of the teams of the competition.

\subsection{The NUS-UoA-UoM Team}
\label{subsec:Mix_Team}
The NUS-UoA-UoM Team is composed by members of the National University of Singapore, of the University of Melbourne, and of the University of Auckland.
\par Based on preliminary experiments, different systems are implemented for each of the different tasks. 
\par For the keystroke task (Task 1), a unimodal system is implemented based only on features pertaining to the keystroke data. Digrams and trigrams are computed to extract typing patterns. The most common ones across enrollment and verification session pairs are selected \cite{10.1145/3386527.3406726}. Then, the set of digrams and trigrams is ordered based on the average time for typing the \textit{n}-gram and the frequency of the \textit{n}-gram in the available text (\textit{n} = 2, 3).  Similarity scores of the two sorted \textit{n}-gram lists are calculated using the \textit{degree of disorder}, following \cite{10.1145/581271.581272}. This value quantifies the similarity between compared sessions. Moreover, from common \textit{n}-grams (with at least 3 occurrences) across enrollment and verification sessions, a \textit{percentage of availability} is computed as an additional feature to capture language similarity. Such quantity is defined as the ratio of the most common \textit{n}-grams included in the list with respect to all \textit{n}-grams. The two computed scores are then averaged. 
\par For the text reading task (Task 2) and for the tapping task (Task 4), only accelerometer and gyroscope data are considered employing rolling time windows of 32 samples to be fed to a separate Long Short-Term Memory (LSTM) Recurrent Neural Network (RNN) \cite{9304922}. For Task 2, a \textit{N}-class classifier is implemented for \textit{N} identities (\textit{N} = 51, the users in the training set). In contrast, for Task 4, a siamese architecture is implemented with triplet loss. A OneClassSVM is trained on the network output embeddings to calculate the similarity between enrolment and verification sessions. Finally, the predictions from the different modalities are fused together using weighted averages and a separate classifier is used for each task. 
\par For the gallery swiping task (Task 3), initially, time window intervals are selected by slicing portions of the touch signals using the touch event information (finger laying and finger lifting). From each of the time windows, the following features are extracted \cite{9367144}:
\begin{itemize}
\itemsep-.5em 
    \item Start and end \textit{(x, y)} position;
    \item Maximum deviation from the mean \textit{x} and mean \textit{y} of all swipe points of each user;
    \item 20$^{th}$, 50$^{th}$ and 80$^{th}$ percentile deviation from the mean \textit{(x, y)} of all swipes of each user;
    \item 20$^{th}$, 50$^{th}$ and 80$^{th}$ percentile of pairwise velocity and acceleration;
    \item Median velocity of the last three points;
    \item Mean acceleration of the first five swipes;
    \item Magnitude sum of pairwise displacements;
    \item Euclidean distance between the start and end \textit{(x, y)} points; 
    \item Ratio of distance and sum of pairwise displacements;
    \item Duration;
    \item Mean velocity.
\end{itemize}
Then, a neural network model composed of four dense layers is trained on-top of these features. The first three layers are composed of 64 units with a SELU activation function, whereas the last one has \textit{N} units. All models are trained for 10000 epochs, with Adam optimizer, and categorical cross-entropy loss function. The model output is used to train a OneClassSVM using enrollment data and that is used for predicting the verification data. 



\subsection{The HCI Essen Team}
\label{subsec:HCI_Essen_Team}
The HCI Essen Team is composed by a member of the University of Duisburg-Essen. 
\par The system implemented is a siamese deep neural network architecture based on a multi-layer perceptron architecture with a contrastive loss function. Siamese Networks have been previously employed within authentication scenarios~\cite{Miller.UsingSiameseNeuralNetworkstoPerformCrossSystemBehavioralAuthenticationinVirtualReality.2021,Zhang.Siameseneuralnetworkbasedgaitrecognitionforhumanidentification.2016}. 
As the acquisition rates of background sensors vary, the number of samples available for each sample is different. Consequently, linear interpolation is used to level out the size of the acquired sequences. Next, a MinMax-algorithm is implemented to normalize the values per acquired sequence. 
\par One model per task has been trained. The different modalities have been fused at data level by stacking all the individual time signals available, and by considering the average duration of the sequences in each of the tasks, applying zero-padding or truncating when necessary.
Paired tuples of the given data are created, where one sample per class is paired with a sample from another class to create input for the contrastive loss. The siamese network implemented consists in a BatchNormalization-layer, followed by four dense layers that come with 400, 200, 100, and 50 units respectively. The activation function for the dense layers is set to the rectified linear unit (ReLU). The Adam optimizer was used with a learning rate of 0.001.
The network is trained for 2000 epochs. 
Although the prediction values of the network are within the interval of $[0, 1)$, they did not cover well the possible distribution. To overcome this issue, they were post-processed by an additional MinMax-normalization.

\subsection{The HBKU CS Lab Team}
\label{subsec:HBKU_CS_Lab_Team}
The HBKU CS Lab Team is composed of members of the Hamad Bin Khalifa University.
\par The same system is implemented for all tasks. The method of Discrete Wavelet Transform (DWT) is applied separately to each time series data of all modalities of each of the tasks. DWT consists in deconstructing a signal into a set of wavelet-basis functions, mutually orthogonal and spatially localized to be non-zero only over certain portions of the signal. The functions are derived from a common function $\phi$, which is called the mother wavelet. Different dilated, scaled and translated versions of this common function $\phi$ are used to generate a set of transforms, each corresponding to a different set of wavelet basis function \cite{Zhang2019}. The Coiflets \cite{sridhar2014wavelet} family of wavelets is used according to the following equation:

\begin{equation}
D[a, b] = \frac{1}{{\sqrt{b}}} \sum_{m=0}^{p-1} f[t_m] \phi \bigg[\frac{t_m-a}{b}\bigg]
\end{equation}
where $a$ represents the translated wavelet across the signal and $b$ represents the compress value, so that a larger value of $b$ represents lower wavelet frequency ($a$ and $b$ are both integers). The mother wavelet is denoted by $\phi$, $p$ is the number of employed functions, and $f[t_m]$ is the discrete input signal \cite{srivastava2018improving}. The DWT is applied as follows:
\begin{equation}
    c_{A_i^j}, c_{B_i^j} = W(D_i^j)
\end{equation}
where $c_A$ and $c_B$ represent the coefficient values after applying wavelet transform in the $i^{th}$ sensor for the $j^{th}$ modality. Therefore, after applying the adopted method, the number of dimensions of the data is doubled. For all modalities apart from the keystroke data, a recursive averaging is performed at modality level to reduce the number of dimensions to their original value. This is done by averaging each set of dimension values with the next, except for the last one and repeating the process until the number of columns is reduced to 3, the original number of dimensions. In the case of the keystroke data, which originally consists in one dimension only, one more dimension is added to the output of the wavelet transform by taking the mean of the two, eventually leading to a three dimensional signal for all modalities. These 3 dimensions are then used as red, green and blue channels to generate an image for each modality. Each task is then represented by a concatenation of all obtained images. Then, a siamese neural network with contrastive loss (based on the on Euclidean distance) is trained to differentiate between the users from the image representation of their patterns. The network is composed of three convolution layers with a varying number of $3\times3$-sized filters (respectively: 64, 128, 256) and ReLU activation function, each of them being followed by a $3\times3$ max pooling operation. The network is trained with the Adam optimizer at a learning rate of 0.00005. The batch size is kept at 32 and the model is trained for 500 epochs. 


\subsection{The JAIRG Team}
\label{subsec:JAIRG_Team}
The JAIRG Team is composed of members of the Jamia Millia Islamia.
\par The submitted models are solely based on touch biometrics for each task. 
For the keystroke task (Task 1), starting from the timestamp information of each key pressed, the first-, second-, and third-order differences are computed as features, together with the corresponding normalized ASCII values.
The text reading task (Task 2) and gallery swiping task (Task 3) data are processed to obtain the following set of features: 
\begin{itemize}
    \itemsep-.5em 
    \item Displacement of swipes;
    \item Velocity of swipes;
    \item Duration of switching between finger laying and finger lifting events, and vice versa; 
    \item Duration of finger laying and finger lifting events.
\end{itemize}
For the tap task (Task 4), the data are pre-processed to obtain the following features: 
\begin{itemize}
    \itemsep-.5em 
    \item Duration of switching between finger laying and finger lifting events, and vice versa;
    \item Duration of finger laying and finger lifting events;
    \item \textit{(x, y)} coordinates.
\end{itemize}
Then, soft Dynamic Time Warping (soft-DTW) values are calculated for genuine, skilled and random forgeries. Finally, such values are used for the training of a Support Vector Machine (SVM), k-Nearest Neighbour (kNN), and Random Forest (RF). The kNN model is trained with 3-neighbors, and based on the Euclidean distance. SVM is trained with a Radial Basis Function kernel. RF with 5 as maximum depth and 100 estimators is trained resulting in the best performance over the validation dataset with respect to the other proposed algorithms.


%% file: 5_Experimental_Results.tex
This section describes the final results of the competition using the evaluation set of BehavePassDB. It should be highlighted that the ranking of MobileB2C is based only on the results achieved during the Evaluation Phase with one winner per task as described in Sec. \ref{subsec:Evaluation_Criteria}. Table \ref{tab:results} shows the results achieved by the participants in each of the tasks in terms of Mixed, Random, and Skilled AUC. Mixed AUC is the only metric valid for the competition final ranking, and it is computed by considering an overall impostor distribution, mixing the random and skilled scenarios. Random and Skilled AUC are used to evaluate the different performances in the two impostor scenarios. 
\par The NUS-UoA-UoM Team is the winner of the task of keystroke (Task 1) with an AUC of 66.37\%. Their system based on the degree of disorder and the percentage of availability of digrams and trigrams outperforms significantly the other teams' approaches based on neural networks (HCI Essen Team), DWT (HBKU CS Lab Team) and soft-DTW (JAIRG Team). It is worth to point out that the system proposed by the NUS-UoA-UoM Team is based on touch features only, in contrast to the approach of the HCI Essen Team and the HBKU CS Lab Team, who exploit background sensor data too.
\par The HCI Essen Team achieves the best performance in the text reading task (Task 2) with an AUC of 57.63\%. In this case, they propose a siamese multilayer perceptron with constrastive loss function based on background sensor data fused at data level. Such architecture appears to be more effective than the multi-class classifier based on a LSTM RNN with categorical cross-entropy loss and OneClassSVM developed by the NUS-UoA-UoM Team for identifying the users.
\par In the gallery swiping task (Task 3), the system based on DWT for the feature extraction and the siamese neural network implemented by the HBKU CS Lab Team proves to be the best performing one, with an AUC of 61.54\%. Moreover, it is interesting to highlight that the systems proposed for this task achieve on average the best results. This could be due to the fact that the horizontal swipes of the gallery swiping task perhaps retain a higher amount of biometric information in order to discriminate among users.
\par Finally, in the tapping task (Task 4), the HBKU CS Lab Team once again have proposed the best performing system. Comparing the results achieved by the proposed systems in all tasks, the task of tapping proves to be the hardest one, possibly due to its shorter duration. 
\par A benchmark of BehavePassDB was carried out in \cite{BehavePassDB} following the same protocol of MobileB2C, based on a separate RNN LSTM network for each of the modalities and fusion at score level, achieving higher results in each of the tasks compared with the ones obtained by the competition participants, thus showing margin for improvement on the provided dataset.

\begin{table}[t!]
\caption{\normalsize The results during the final Evaluation Phase in terms of Mixed, Random, and Skilled AUC [\%] are displayed below.}
\label{tab:results}
\resizebox{.45\textwidth}{!}{
\begin{tabular}{ccccc}
\hline
\multicolumn{1}{|c|}{\textbf{\#}} & \multicolumn{1}{c|}{\textbf{\begin{tabular}[c]{@{}c@{}}Team\\ Name\end{tabular}}} & \multicolumn{1}{c|}{\textbf{\begin{tabular}[c]{@{}c@{}}Mixed\\ AUC {[}\%{]}*\end{tabular}}} & \multicolumn{1}{c|}{\textbf{\begin{tabular}[c]{@{}c@{}}Random\\ AUC {[}\%{]}\end{tabular}}} & \multicolumn{1}{c|}{\textbf{\begin{tabular}[c]{@{}c@{}}Skilled\\ AUC {[}\%{]}\end{tabular}}} \\ \hline
\multicolumn{5}{|c|}{\textbf{Task 1 - Keystroke}}                                                                                                                                                                                                                                                                                                                                                                \\ \hline
\multicolumn{1}{|c|}{\textbf{1}}  & \multicolumn{1}{c|}{\textbf{NUS-UoA-UoM}}                                                   & \multicolumn{1}{c|}{\textbf{66.37}}                                                         & \multicolumn{1}{c|}{\textbf{64.77}}                                                         & \multicolumn{1}{c|}{\textbf{67.91}}                                                          \\ \hline
\multicolumn{1}{|c|}{2}           & \multicolumn{1}{c|}{HCI Essen}                                                            & \multicolumn{1}{c|}{51.12}                                                                  & \multicolumn{1}{c|}{53.02}                                                                  & \multicolumn{1}{c|}{51.23}                                                                   \\ \hline
\multicolumn{1}{|c|}{3}           & \multicolumn{1}{c|}{HBKU CS Lab}                                                            & \multicolumn{1}{c|}{51.25}                                                                  & \multicolumn{1}{c|}{49.38}                                                                  & \multicolumn{1}{c|}{53.13}                                                                   \\ \hline
\multicolumn{1}{|c|}{4}           & \multicolumn{1}{c|}{JAIRG}                                                            & \multicolumn{1}{c|}{45.57}                                                                  & \multicolumn{1}{c|}{52.29}                                                                  & \multicolumn{1}{c|}{39.89}                                                                   \\ \hline
\multicolumn{5}{|c|}{\textbf{Task 2 - Text Reading}}                                                                                                                                                                                                                                                                                                                                                             \\ \hline
\multicolumn{1}{|c|}{\textbf{1}}  & \multicolumn{1}{c|}{\textbf{HCI Essen}}                                                   & \multicolumn{1}{c|}{\textbf{57.63}}                                                         & \multicolumn{1}{c|}{\textbf{61.27}}                                                         & \multicolumn{1}{c|}{\textbf{53.98}}                                                          \\ \hline
\multicolumn{1}{|c|}{2}           & \multicolumn{1}{c|}{NUS-UoA-UoM}                                                            & \multicolumn{1}{c|}{54.89}                                                                  & \multicolumn{1}{c|}{58.49}                                                                  & \multicolumn{1}{c|}{51.29}                                                                   \\ \hline
\multicolumn{1}{|c|}{3}           & \multicolumn{1}{c|}{JAIRG}                                                            & \multicolumn{1}{c|}{50.63}                                                                  & \multicolumn{1}{c|}{50.00}                                                                  & \multicolumn{1}{c|}{51.25}                                                                   \\ \hline
\multicolumn{1}{|c|}{4}           & \multicolumn{1}{c|}{HBKU CS Lab}                                                            & \multicolumn{1}{c|}{48.27}                                                                  & \multicolumn{1}{c|}{59.42}                                                                  & \multicolumn{1}{c|}{37.13}                                                                   \\ \hline
\multicolumn{5}{|c|}{\textbf{Task 3 - Gallery Swiping}}                                                                                                                                                                                                                                                                                                                                                          \\ \hline
\multicolumn{1}{|c|}{\textbf{1}}  & \multicolumn{1}{c|}{\textbf{HBKU CS Lab}}                                                   & \multicolumn{1}{c|}{\textbf{61.54}}                                                         & \multicolumn{1}{c|}{\textbf{67.35}}                                                         & \multicolumn{1}{c|}{\textbf{55.73}}                                                          \\ \hline
\multicolumn{1}{|c|}{2}           & \multicolumn{1}{c|}{JAIRG}                                                            & \multicolumn{1}{c|}{55.94}                                                                  & \multicolumn{1}{c|}{56.25}                                                                  & \multicolumn{1}{c|}{55.63}                                                                   \\ \hline
\multicolumn{1}{|c|}{3}           & \multicolumn{1}{c|}{NUS-UoA-UoM}                                                            & \multicolumn{1}{c|}{55.66}                                                                  & \multicolumn{1}{c|}{55.54}                                                                  & \multicolumn{1}{c|}{55.77}                                                                   \\ \hline
\multicolumn{1}{|c|}{4}           & \multicolumn{1}{c|}{HCI Essen}                                                            & \multicolumn{1}{c|}{54.72}                                                                  & \multicolumn{1}{c|}{57.30}                                                                  & \multicolumn{1}{c|}{52.14}                                                                   \\ \hline
\multicolumn{5}{|c|}{\textbf{Task 4 - Tapping}}                                                                                                                                                                                                                                                                                                                                                                  \\ \hline
\multicolumn{1}{|c|}{\textbf{1}}  & \multicolumn{1}{c|}{\textbf{HBKU CS Lab}}                                                   & \multicolumn{1}{c|}{\textbf{59.58}}                                                         & \multicolumn{1}{c|}{\textbf{57.22}}                                                         & \multicolumn{1}{c|}{\textbf{61.94}}                                                          \\ \hline
\multicolumn{1}{|c|}{2}           & \multicolumn{1}{c|}{NUS-UoA-UoM}                                                            & \multicolumn{1}{c|}{52.39}                                                                  & \multicolumn{1}{c|}{54.72}                                                                  & \multicolumn{1}{c|}{50.06}                                                                   \\ \hline
\multicolumn{1}{|c|}{3}           & \multicolumn{1}{c|}{JAIRG}                                                            & \multicolumn{1}{c|}{46.25}                                                                  & \multicolumn{1}{c|}{48.75}                                                                  & \multicolumn{1}{c|}{43.75}                                                                   \\ \hline
\multicolumn{1}{|c|}{4}           & \multicolumn{1}{c|}{HCI Essen}                                                            & \multicolumn{1}{c|}{43.89}                                                                  & \multicolumn{1}{c|}{40.16}                                                                  & \multicolumn{1}{c|}{47.62}                                                                   \\ \hline
\multicolumn{5}{l}{*Valid for the final ranking of MobileB2C.}                                                                                                                                                                                                                                                                          
\end{tabular}}
\end{table}

\subsection{Random vs. Skilled Impostor Analysis}
\label{Random_vs_Skilled_Impostor_Analysis}
The results shown in the fourth and fifth columns of Table \ref{tab:results}, while not valid for the competition final ranking, are useful to compare the system performance in each of the impostor scenarios. Around 62\% of the times the systems proposed perform better in the random impostor case, achieving on average +1.70\% AUC in absolute terms. This trend shows that in this experimental set up the skilled scenario represents a more difficult case, and that some learning bias is introduced by the device due to sensor differences and calibration imperfections. In any case, it should not be overlooked the fact that the training set only contains random impostor forgeries as the data of each of the subjects have been acquired on a different device. In light of this, the models could be optimized for this scenario. Including skilled forgeries in the training process could be beneficial, however the data acquisition can be difficult to carry out rigorously and on a large scale for the skilled impostor scenario.

%% file: 6_Conclusions.tex
This paper has described the experimental framework and results of the IJCB 2022 Mobile Behavioral Biometrics Competition (MobileB2C). The goal of MobileB2C is carrying out a benchmark evaluation of the latest state-of-the-art mobile behavioral biometrics authentication technology using a public database, BehavePassDB. MobileB2C is based on four different tasks, corresponding to common mobile device use cases: \textit{(i)} texting, \textit{(ii)} text reading, \textit{(iii)} gallery swiping, \textit{(iv)} tapping. In each of the tasks, the touchscreen information is recorded along with the data of five background sensors. The database used encompasses both traditional “random impostor” scenario (different users with different devices), but also the challenging “skilled impostor” scenario (different users on the same device attempting to imitate the device owner). 
\par The results achieved in the final Evaluation Phase of MobileB2C show the feasibility of user authentication through behavioral biometrics, although this proves to be a non-trivial challenge. The NUS-UoA-UoM Team is the winner of the task of keystroke (Task 1) with a Mixed AUC of 66.37\%, exploiting a unimodal system based on touch information only. The HCI Essen Team proposes the best performing algorithm for Task 2, text reading (Mixed AUC of 57.63\%), based on a multi-layer perceptron architecture with a contrastive loss function, and fusion at data level. Finally, the HBKU CS Lab Team wins the gallery swiping and tapping tasks, Task 3 and Task 4 (Mixed AUC values respectively 61.54\% and 59.58\%), by proposing a system based on DWT and a siamese neural network. In most cases, the skilled impostor scenario proves to be harder, leading to the conclusion that a non-negligible device bias is learned.
\par MobileB2C will be established as an on-going competition \footnote{https://sites.google.com/view/mobileb2c/}, where researchers can easily benchmark their systems against the state of the art in an open common platform using a public database, BehavePassDB.